\documentclass[10pt,twocolumn,letterpaper]{article}
\usepackage{duckuments}

\usepackage[review]{cvpr}      
\usepackage{color}

\usepackage{tabularx, booktabs}
\newcolumntype{Y}{>{\centering\arraybackslash}X}
\usepackage{makecell}
\usepackage{threeparttable}
\usepackage{hyperref}
\usepackage{marvosym}

\definecolor{cvprblue}{rgb}{0.21,0.49,0.74}
\usepackage{hyperref}
\hypersetup{hypertex=true,
    colorlinks=true,
    linkcolor=cvprblue,
    anchorcolor=cvprblue,
    citecolor=cvprblue}
\hypersetup{pdfborder={0 0 0}}

\title{3DSceneEditor: Controllable 3D Scene Editing with Gaussian Splatting}

\author{Ziyang Yan\\
3D Optical Metrology unit, Fondazione Bruno Kessler\\
Via Sommarive, 18, 38123 Povo, Trento, Italy\\
Department of Information Engineering and Computer Science, University of Trento\\
Via Sommarive, 9, 38123 Povo, Trento, Italy\\
{\tt\small zyan@fbk.eu}
\and
Second Author\\
Institution2\\
First line of institution2 address\\
{\tt\small secondauthor@i2.org}
\and
Yihua Shao\\
Institution2\\
First line of institution2 address\\
{\tt\small secondauthor@i2.org}
\and
Fabio Remondino\\
3D Optical Metrology unit, Fondazione Bruno Kessle\\
Via Sommarive, 18, 38123 Povo, Trento, Italy\\
{\tt\small remondino@fbk.eu}
}

\makeatletter
\def\thanks#1{\protected@xdef\@thanks{\@thanks
        \protect\footnotetext{#1}}}
\makeatother

\begin{document}

\maketitle

\begin{abstract} 
The creation of 3D scenes has traditionally been both labor-intensive and costly, requiring designers to meticulously configure 3D assets and environments. Recent advancements in generative AI, including text-to-3D and image-to-3D methods, have dramatically reduced the complexity and cost of this process. However, current techniques for editing complex 3D scenes continue to rely on generally interactive multi-step, 2D-to-3D projection methods and diffusion-based techniques, which often lack precision in control and hamper real-time performance. In this work, we propose \textbf{3DSceneEditor}, a fully 3D-based paradigm for real-time, precise editing of intricate 3D scenes using Gaussian Splatting. Unlike conventional methods, 3DSceneEditor operates through a streamlined 3D pipeline, enabling direct manipulation of Gaussians for efficient, high-quality edits based on input prompts. The proposed framework (i) integrates a pre-trained instance segmentation model for semantic labeling; (ii) employs a zero-shot grounding approach with CLIP to align target objects with user prompts; and (iii) applies scene modifications, such as object addition, repositioning, recoloring, replacing, and deletion—directly on Gaussians. Extensive experimental results show that 3DSceneEditor achieves superior editing precision and speed with respect to current SOTA 3D scene editing approaches, establishing a new benchmark for efficient and interactive 3D scene customization.

\end{abstract}    
\section{Introduction}
\label{sec:intro}

The creation and editing of 3D scenes have been both costly and time-consuming. Designers have to manually work with various 3D tools, investing considerable time and effort in tasks like sketching, designing layouts, arranging objects, and selecting material textures \cite{wang2024gaussianeditor}. However, the recent emergence of generative AI has revolutionized these processes, allowing the creation of high-quality 3D assets faster and more affordable. Using text-to-3D \cite{poole2022dreamfusion, huang2024dreamcontrol, li2024instant3d, liu2024sherpa3d, chen2023control3d, raj2023dreambooth3d, lin2023magic3d,liu2024pi3d, wu2025tpa3d, chen2024text,li2023instant3d, yu2023points, yi2024gaussiandreamerpro, yi2023gaussiandreamer} or image-to-3D \cite{liu2024isotropic3d, liu2024one, hong2023lrm, liu2024one2, long2024wonder3d, ye2024consistent, liu2023zero, tang2023dreamgaussian, tang2023make} methods, users can now quickly generate or re-layout detailed 3D scenes using text prompts or images. As a result, AI-driven generation techniques have gradually gained widespread popularity across industries like advertising, animation, game development, and VR/AR.

Before the development of Gaussian Splatting \cite{kerbl20233d}, NeRF-based methods \cite{sun2024nerfeditor, lazova2023control, dong2024vica, jambon2023nerfshop, song2023blending, bao2023sine, yuan2022nerf} dominate the field of 3D editing due to their powerful 3D scene representation capabilities \cite{yan2023nerfbk,remondino2023critical}. These methods typically rely on pre-trained NeRF models to edit 3D scenes. A notable example is Instruct-NeRF2NeRF \cite{haque2023instruct}, which uses an image-conditional 2D diffusion model called InstructPix2Pix \cite{brooks2023instructpix2pix}, for 3D scene editing. However, NeRF’s dependence on high-dimensional multilayer perceptron (MLP) networks for encoding scene data limits its ability to directly modify specific scene elements and complicated tasks such as inpainting and scene composition \cite{chen2024gaussianeditor}. Additionally, NeRF’s implicit representation and resource-demanding issues pose a significant challenge for real-time editing.

The emergence of 3D Gaussian Splatting has revolutionized both 3D reconstruction and image rendering, with significant impacts on 3D editing. 3D Gaussian Splatting (3D-GS) \cite{kerbl20233d} is a pioneering technique that achieves real-time rendering while maintaining high-quality outputs with fast training speed. Its explicit representation offers distinct advantages for editing, as each 3D Gaussian is individually manipulable, allowing for direct and efficient scene modifications. This innovation has inspired the creation of several 3D editing methods based on Gaussian Splatting, such as Instruct-GS2GS \cite{igs2gs}, GaussianEditor \cite{chen2024gaussianeditor, wang2024gaussianeditor}, etc., which are built on InstructPix2Pix \cite{brooks2023instructpix2pix}. However, these editing approaches, fully based on diffusion models, often lack detailed control over scene modifications and are limited by input image resolution. For example, InstructPix2Pix, built on stable diffusion \cite{rombach2022high}, primarily supports 512x512 or 768x768 px images, and deviations from these resolutions can significantly impact the quality of the output \cite{moreno2023analysis}.

Since controllable scene editing in complex layouts using pre-trained generative models is highly challenging, current methods \cite{ye2023gaussian, cen2023saga, shen2025flashsplat, choi2024click, hu2024semantic}, largely based on Gaussian Splatting, rely on models like Grounding Dino \cite{liu2023grounding} and SAM \cite{kirillov2023segment} to detect and segment objects in each 2D image before projecting features into 3D space. This 2D-based approach complicates the process, requiring users to render multiple 2D images from a 3D model, segment and ground each object, and project them frame-by-frame into 3D space. Therefore, achieving real-time, user-friendly, and controllable editing would mark a significant breakthrough.

In this paper, we propose an editing paradigm shift with a 3D-only approach, named \textbf{3DSceneEditor}, for controllable editing of complex scenes based on Gaussian Splatting. Input only one prompt, 3DSceneEditor can achieve precise edits within seconds. The key to achieve real-time editing is our fully 3D pipeline, which allows direct manipulation of Gaussians in a single step. We first use a pre-trained instance segmentation model from Mask3D \cite{schult2023mask3d} to assign semantic labels to each Gaussian. Next, we ground target instances using a zero-shot grounding algorithm and employ CLIP \cite{radford2021learning} to align target objects with the input prompt and the desired edits. Finally, the editing operations are applied directly to the Gaussians, and the entire process can be completed in just tens of seconds. To handle potential mis-segmentations from Mask3D, we use KNN to correct outliers through voting. Experimental results show that our pipeline outperforms current SOTA approaches in editing quality, processing time and GPU usage.

In summary, the primary contributions of the paper are:

\begin{itemize}

\item [1)] 
a 3D-only editing approach, named \textbf{3DSceneEditor}, for complex indoor scenes: Unlike previous multi-step methods that always rely on 2D-3D semantic projection, our framework simplifies the process, enabling real-time editing, higher quality results, and improved user interaction.
\item [2)] 
an innovative zero-shot instance grounding pipeline for precise grounding of target objects in complex 3D layouts, which is achieved through prompt-based keyword extraction, view-based relationships simplified with a 2D egocentric approach, and language-object correlation using a multimodal language model.
\item [3)] 
A controllable scene editing method enabling object addition, movement, recoloring, removal, and replacement through text-based instructions, using a 3D Gaussian-based model for efficient 3D scene reconstruction and direct manipulation.

\end{itemize}


\section{Related Work}
\label{sec:related work}

\subsection{3D Representations}
Neural Radiance Field (NeRF) \cite{mildenhall2021nerf}, based on implicit representation and volumetric rendering, has been a representative work in the field of 3D reconstruction in recent years. It has been widely used for 3D reconstruction \cite{tancik2022block, zhong2025structured, wang2022neuris, chen2023structnerf, ni2024phyrecon, liu2024neuralsurfacereconstructionrendering}, AI generation \cite{metzer2023latent, huang2024dreamcontrol, chen2023single}, and 3D editing \cite{yuan2022nerf, lazova2023control, he2024customize}. However, NeRF-based models require dense and continuous sampling in 3D space for optimization. When dealing with complex scenes like ScanNet \cite{dai2017scannet} or ScanNet++ \cite{yeshwanth2023scannet++} (each scene with hundreds or even thousands of images), the relatively long training time, high computational demand, and substantial GPU memory requirements reduce user friendliness and make real-time scene editing challenging. Recently, 3D Gaussian Splatting (3D-GS) \cite{kerbl20233d} has become the leading 3D representation technique, praised for its quick training time and high-quality real-time rendering. Similar to NeRF, beside 3D reconstruction purposes \cite{yan2024renderworld, lu2024scaffold, huang2024sc, huang20242d, zhu2025fsgs}, 3D-GS is also being widely adopted for 3D generation \cite{tang2023dreamgaussian, yi2023gaussiandreamer, yi2024gaussiandreamer, DiffGS} and editing \cite{chen2024gaussianeditor, wang2024gaussianeditor, shen2025flashsplat, cen2023saga, hu2024semantic, hsu2024autovfx, progresseditor}. Considering our work is mainly working on editing complex indoor scenes, we apply 3D-GS as a 3D representation method, and it can significantly accelerate the 3D editing process.
\subsection{3D Scene Editing}
Editing NeRF is inherently challenging due to the complex interplay between shape and appearance \cite{chen2024gaussianeditor}. However, the availability to individually edit each Gaussian in 3D-GS provides significant flexibility for scene editing, particularly in indoor environments with intricate layouts. 
Existing 3D-GS editing methods fall into two main categories. The first relies on 2D diffusion priors or large language models (LLMs) \cite{chen2024gaussianeditor, wang2024gaussianeditor, igs2gs, chen2024dge, xiao2024localized, zhang20243DitScene}, enabling text-driven editing pipelines but often limited in complex scenes by diffusion model capabilities. The second category directly edits 3D scenes, bypassing diffusion models by using 2D masks generated by models like Segment Anything Model (SAM) \cite{kirillov2023segment} and Ground SAM \cite{ren2024grounded} for 2D-3D semantic projection. For instance, Gaussian Grouping \cite{ye2023gaussian} enhances precision by tracking objects across frames, while SAGA \cite{cen2023saga} and SAGD \cite{hu2024semantic} assign 3D semantic features through 2D mask projections. FlashSplat \cite{shen2025flashsplat} simplifies this by treating 2D mask lifting as a linear programming problem, enabling single-step editing. However, these methods lack prompt-based control, requiring manual scene edits, which limits usability. 

In contrast, our 3D-only framework directly interprets scenes in 3D, assigning semantic labels to each Gaussian via a pre-trained 3D segmentation module. This streamlined approach removes projection overhead, enables real-time editing within seconds, and overcomes the limitations of using 2D diffusion models. Additionally, our integrated Open Vocabulary module enhances intuitiveness and user-friendliness.
%


\section{Methodology}


\begin{figure*}[t]
   \centering
   \includegraphics[width=\linewidth]{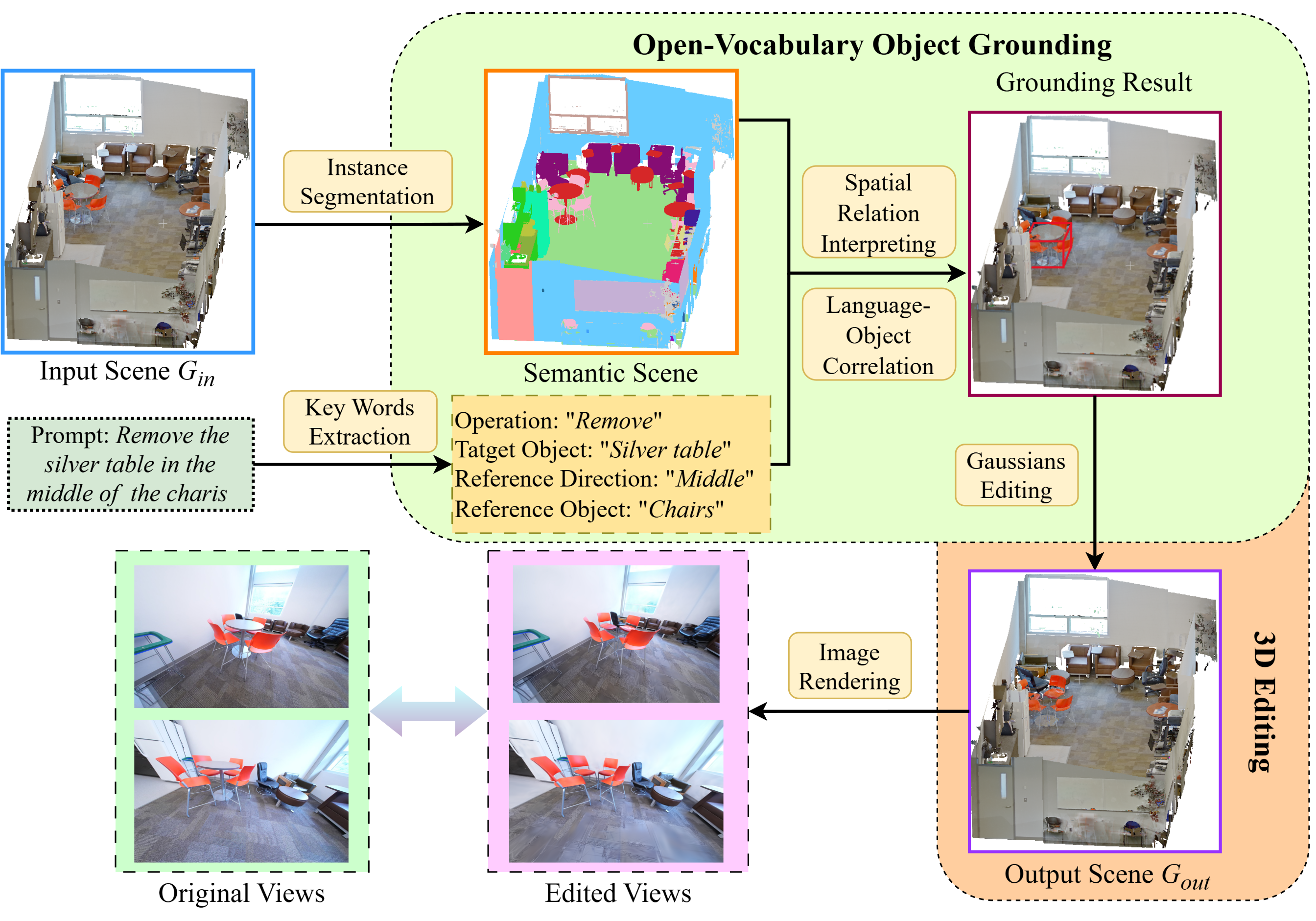}
   \caption{Our paradigm, named \textbf{3DSceneEditor}, consists of three key steps. First, a pre-trained instance segmentation model is applied to understand the input scene and assign a semantic label to each Gaussian. Followed by an Open Vocabulary Object Grounding module, which is used to ground the target objects from the input semantic Gaussians and generate the ROI for target objects. Finally, we execute the specified scene editing operation in ROI based on the prompt and render the edited views.}
   \label{fig:pipeline}
   \vspace{-1.0em}
 \end{figure*}

First, we provide an overview of our proposed 3D-only approach (Section \ref{section:3.1}), followed by an introduction to our Open-Vocabulary Object Grounding module, which uses a view-dependent module and multimodal alignment assistant (CLIP) (Section \ref{section:3.2}). Finally, Section \ref{section:3.3} covers the 3D editing operations and optimization module.



\subsection{Overall Framework}
\label{section:3.1}
3D Gaussian Splatting \cite{kerbl20233d} is an innovative approach that represents a 3D scene explicitly as a set of Gaussians $\left \{{{G_x}}\right \}_{x=1}^N$. Our editing pipeline (Fig. \ref{fig:pipeline}) starts from a set of 3D Gaussians ${G_{in}}$ trained from a specific scene and a prompt $\tau$. These Gaussians are processed through a pretrained instance segmentation model that assigns semantic labels to each Gaussian. Next, target objects and their references are identified based on the keywords extracted from the prompt, and their Region of Interest (ROI) is determined using an Open-Vocabulary Object Grounding module (Section \ref{section:3.2}). With this information and editing directives from the prompt, our pipeline enables real-time scene editing through direct manipulations of Gaussians within the ROI, supporting operations such as object addition, movement, removal, replacement, and colorization (Section \ref{section:3.3}). Thus, our editing task can be defined as:
\begin{equation}
G_{out} =Edit({G_{in}}, \tau).
\end{equation}

\subsection{Open-Vocabulary Object Grounding}
\label{section:3.2}
\begin{figure}[h]
   \centering
   \includegraphics[width=\linewidth]{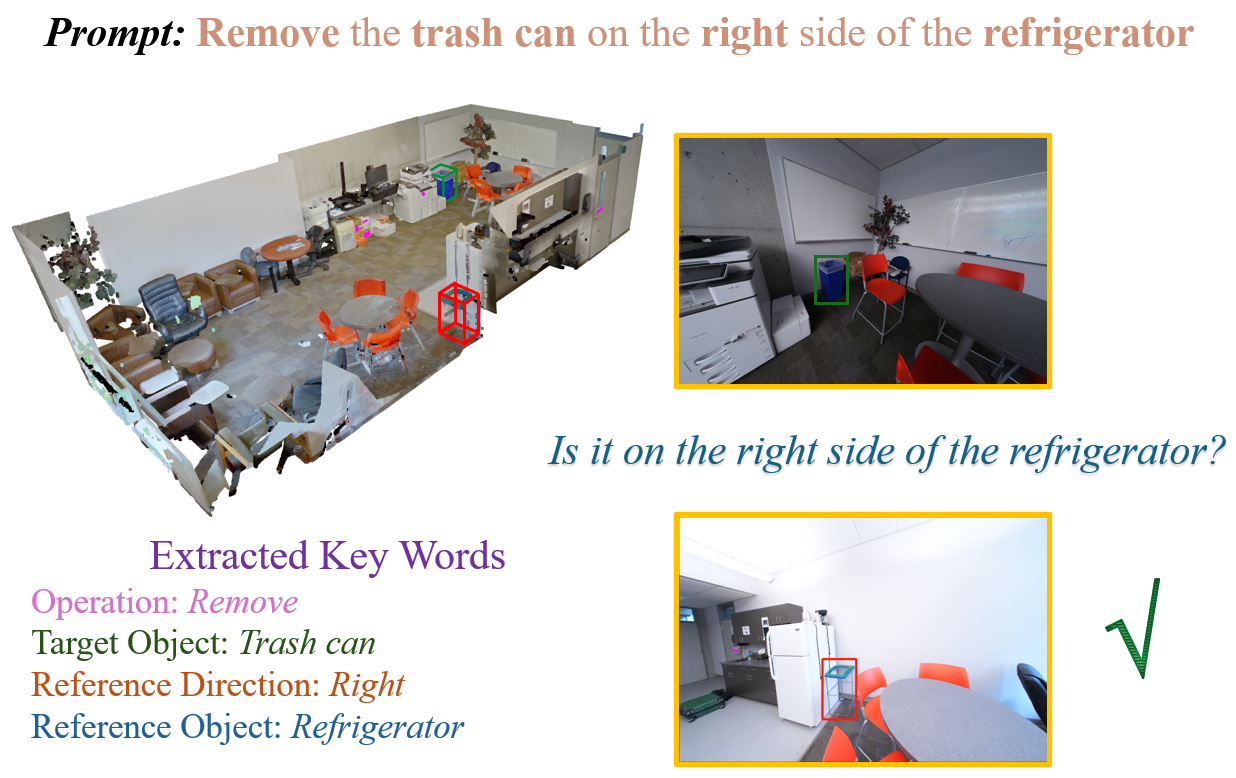}
   \caption{\textbf{Visualization of our Object Grounding.} We first extract the key words from the prompt (\textbf{bold fonts} in the picture). Since the positional relationships between objects in 3D space change in different viewpoints, we need to project them onto a static 2D plane to better understand the scene.}
   \label{fig:Object Groudning}
 \end{figure}
 
3D Objects Grounding is an essential part of our 3D-only editing pipeline. Prior approaches, like Gaussian grouping \cite{ye2023gaussian} and FlashSplat \cite{shen2025flashsplat}, extract frame-level features using pre-trained 2D detection and segmentation models, then project semantic labels from 2D images into 3D space, dividing Gaussians into instance-based groups. However, these methods neglect the inherent spatial relationships within 3D Gaussians, posing major limitations for complex editing tasks. For instance, similar or identical objects are common in 3D scenes, especially indoors. Even though groups of Gaussians allow for object removal or recoloring by directly operating on Gaussian clusters, performing more complex interactions, such as adding objects or swapping their positions, remains extremely challenging. Additionally, these pipelines require users to know in advance which instances each Gaussian group represents, rather than enabling direct localization of target objects through interactive prompts. To address these challenges, we introduce an open-vocabulary 3D grounding module as shown in Fig. \ref{fig:Object Groudning}.

\textbf{Key words extraction.} For editing a given 3D scene, we first build a specialized vocabulary set, which includes query terms for various instances (e.g., ”coffee table," "monitor") and scene-editing keywords (e.g., "remove," ”recolor”), view-dependents (e.g., "left" and "between"), and color mappings to capture specific keywords from the prompt and classify them into "Operation," "Target Object", "Reference Direction" and "Reference Object". Based on the semantics of the Gaussians, we then filter and identify candidate objects that meet the specified categories.

\textbf{Spatial relation interpreting.} In complex 3D scenes, distinguishing objects of the same category presents a significant challenge. Our 3D grounding module addresses this by interpreting spatial relations for each candidate object. We employ the 2D egocentric, view-related module introduced by \cite{yuan2024visual} to simulate a camera at the center of the scene, then project the complex geometric relationships between target and reference objects from 3D space onto a 2D plane to enable pixel-level filtering of candidate objects based on view-dependent relations.

\textbf{Language-Object correlation} Finally, an Image-Text Alignment module is applied \cite{radford2021learning} to evaluate the cosine similarity between the prompt and the tokenized image query, finding the optimal candidate target objects and returning their 3D bounding boxes as ROI, which is curial for the following 3D Gaussian editing.

\subsection{3D Gaussian Editing}
\label{section:3.3}
The proposed pipeline initiates the 3D editing operation based on prompt instructions, applying editing only to the Gaussians located within the specified ROI, as reported in Section \ref{section:3.3}. It can support totally 5 types operations: object removal, re-colorization, object addition, object movement and object replacement.

\textbf{Object removal and re-coloration.} 
Our approach can easily achieve object removal and re-coloration by either removing Gaussians or directly changing their color feature with the target semantic labels. To facilitate prompt-based re-colorization, we first construct a color mapping table with common colors and map the color keyword from prompt to color table directly.

\begin{figure}[h]
   \centering
   \includegraphics[width=\linewidth]{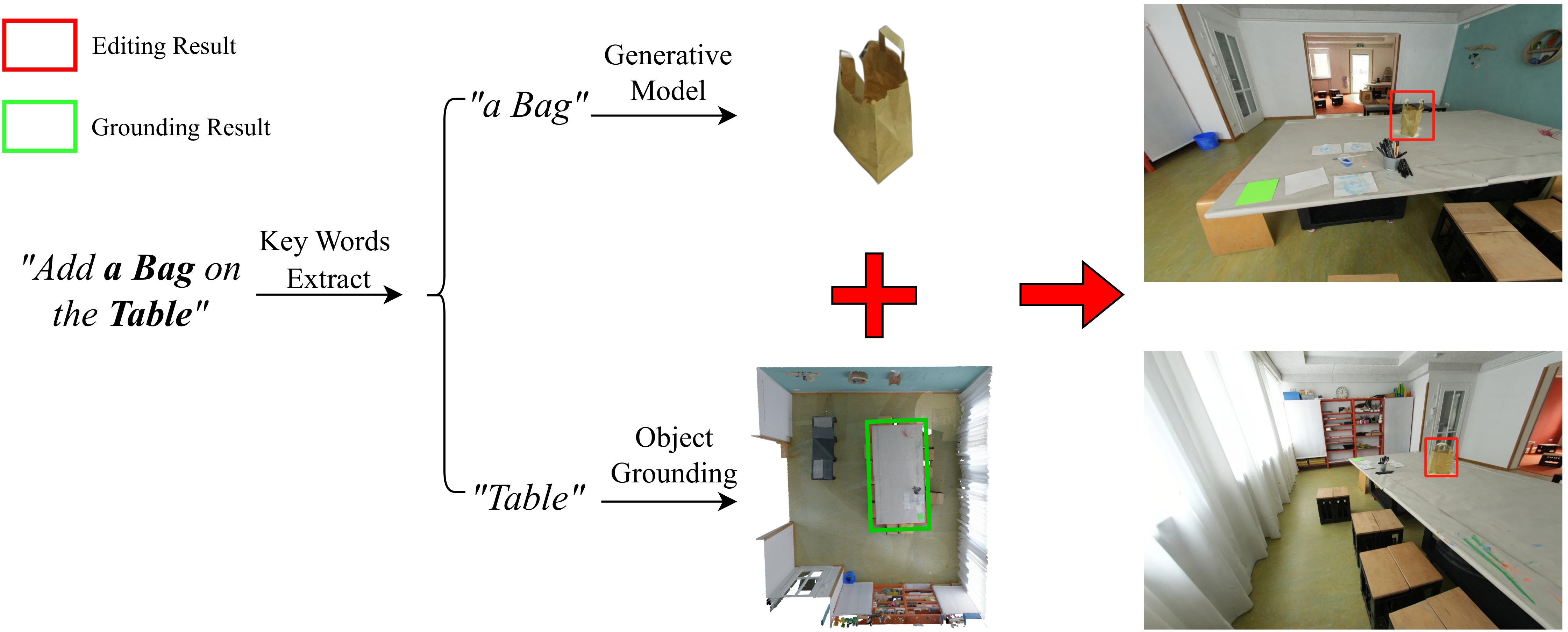}
   \caption{\textbf{Visualization of our object addition pipeline.} We generate new objects using a Gaussian-based generative model, guided by keywords extracted from the prompt. With the assistance of the Object Grounding module, these new Gaussians are then integrated into the ROI within the input scene.}
   \label{fig:pipeline_addition}
    \vspace{-1.0em}
 \end{figure}

\textbf{Object addition and replacement.} 
Other image-based methods \cite{wu2024gaussctrl, zhang20243DitScene, chen2024gaussianeditor} primarily rely on 2D diffusion priors and novel view synthesis to add or replace objects in scenes. In contrast, our pipeline achieves object integration directly in 3D space by generating new objects from prompts or images using a Gaussian-based generative model. We then incorporate these objects by adding the new Gaussians or substituting them within ROI, as shown in Fig. \ref{fig:pipeline_addition}.

Since the size of AI-generated objects can be unpredictable, we first apply an adjustable scaling parameter to match their size to reference objects. We then position the Gaussians of new objects into target regions or replace existing objects with the newly generated Gaussians. For object addition, our method aligns the central axes of the external bounding boxes of both the new and reference objects, ensuring their corresponding bounding box sides overlap based on the target view-dependent relationship.

This geometry-based stitching technique effectively minimizes prediction errors, commonly encountered in diffusion-prior-based methods, making it highly applicable to a wide range of 3D scenes with complex layouts.

\textbf{Object movement.} 
To achieve object movement, we select the valid Gaussians from their semantic labels and lightly adjust their coordinates ($x_in, y_in, z_in$) in the world coordinate system based on their reference objects and text instruction ("close," "far away"). Moving 3D Gaussians are inherently complicated. Thus, each Gaussian not only represents one object attribute \cite{shen2025flashsplat}. Since each 3D Gaussian is projected onto a 2D plane via orthographic projection, the Gaussian’s covariance in ray space is derived by applying a series of transformations to the Gaussian's covariance matrix $\Sigma$ and its center ($x_0, y_0, z_0$) in world coordinate. Moving a 3D Gaussian can affect other objects along the same ray in ray space, resulting in extra noise and artifacts in the projected 2D image, which impact different objects across various viewpoints. As a result, our pipeline currently only supports moving small objects within a limited range to avoid displacing too many Gaussians at once, which could disrupt scene rendering significantly.

\textbf{Optimization of editing.}
Since pre-trained instance segmentation models may not perform well in certain specialized scenarios (e.g., objects positioned near the junction between walls and floors), we pre-process the scene by applying K-Nearest Neighbors (KNN) clustering to re-label the Gaussians within ROI before editing. For artifacts or “black holes” that may appear in the background after object removal, we apply KNN again and perform inpainting based on the Gaussian features of the nearest background points. Our ablation study (Section \ref{Optimization of Inpainting}) validates the necessity of this editing optimization module.
\section{Experiment}
\subsection{Implementation Details}
Our method is implemented in PyTorch and CUDA, with all 3D Gaussians trained using the original 3D Gaussian Splatting \cite{kerbl20233d} and DreamGaussian \cite{tang2023dreamgaussian}. Experiments were conducted on a single NVIDIA Tesla A100 GPU using 11 representative indoor scenes from the ScanNet++ \cite{yeshwanth2023scannet++} dataset, including kindergartens, offices, rest rooms and studios, with prompts customized for each scene layout. Since ScanNet++ images are captured by a fisheye digital SLR camera, which is incompatible with 3D Gaussian Splatting, we utilize the ScanNet++ Toolkit \cite{yeshwanth2023scannet++} to undistort the fisheye images and convert the camera model to a pinhole model with COLMAP \cite{schonberger2016structure}. To support a variety of editing applications, we use a pre-trained instance segmentation model from ScanNet200 \cite{rozenberszki2022language} to obtain Gaussian semantics.

\subsection{Qualitative Evaluation}
\begin{figure*}[h]
   \centering
   \includegraphics[width=\linewidth]{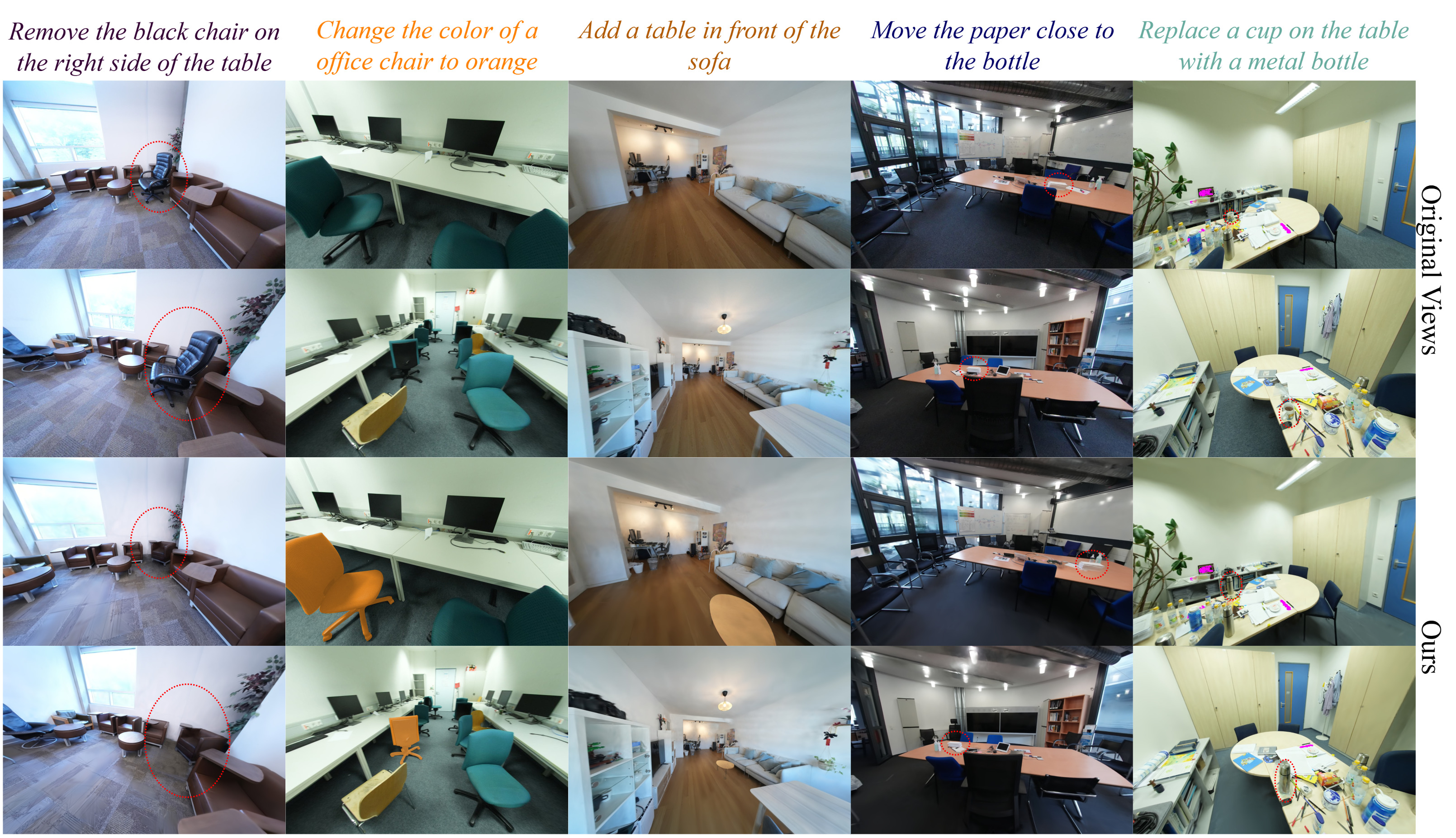}
   \caption{\textbf{Extensive Results of 3DSceneEditor.} This figure presents additional results across diverse scenes, demonstrating that our method enables precise and varied scene editing for layouts and objects of different scales.}
   \label{fig:Extensive Results}
    \vspace{-1.0em}
 \end{figure*}
 
\textbf{Visualization results of different scenes.} Fig. \ref{motivation} and Fig. \ref{fig:Extensive Results} present visual results from 3DSceneEditor, demonstrating its capability for precise, controllable, and 3D-consistent editing. In Fig. \ref{motivation}, our pipeline represents various editing operations on individual objects within a 3D scene. The left side of the figure shows original scenes and the object grounding results, where the stool is anchored to the left of the table. The right side illustrates different edits applied to the stool using varied text instructions. By leveraging Gaussian 3D spatial information and memorizing ROI, our method minimizes the resources consumed in repeated semantic reasoning and grounding of the same scene, significantly reducing secondary edit time to just seconds. Fig. \ref{fig:Extensive Results} further showcases these capabilities across diverse 3D scenes. In the first two columns, we demonstrate object removal (a black chair) and color modification (an office chair). Despite multiple identical chairs in these scenes, our pipeline accurately grounds and edits objects based on prompts. While the middle column showcases our method’s precision in adding objects via interacting with generative models. The fourth and fifth columns highlight movement and replacement of small objects within complex scenes; even with numerous small items, our approach maintains precise object grounding and editing, delivering robust editing capabilities across complex scenes and varied object types.

\begin{figure*}[h]
   \centering
   \includegraphics[width=\linewidth]{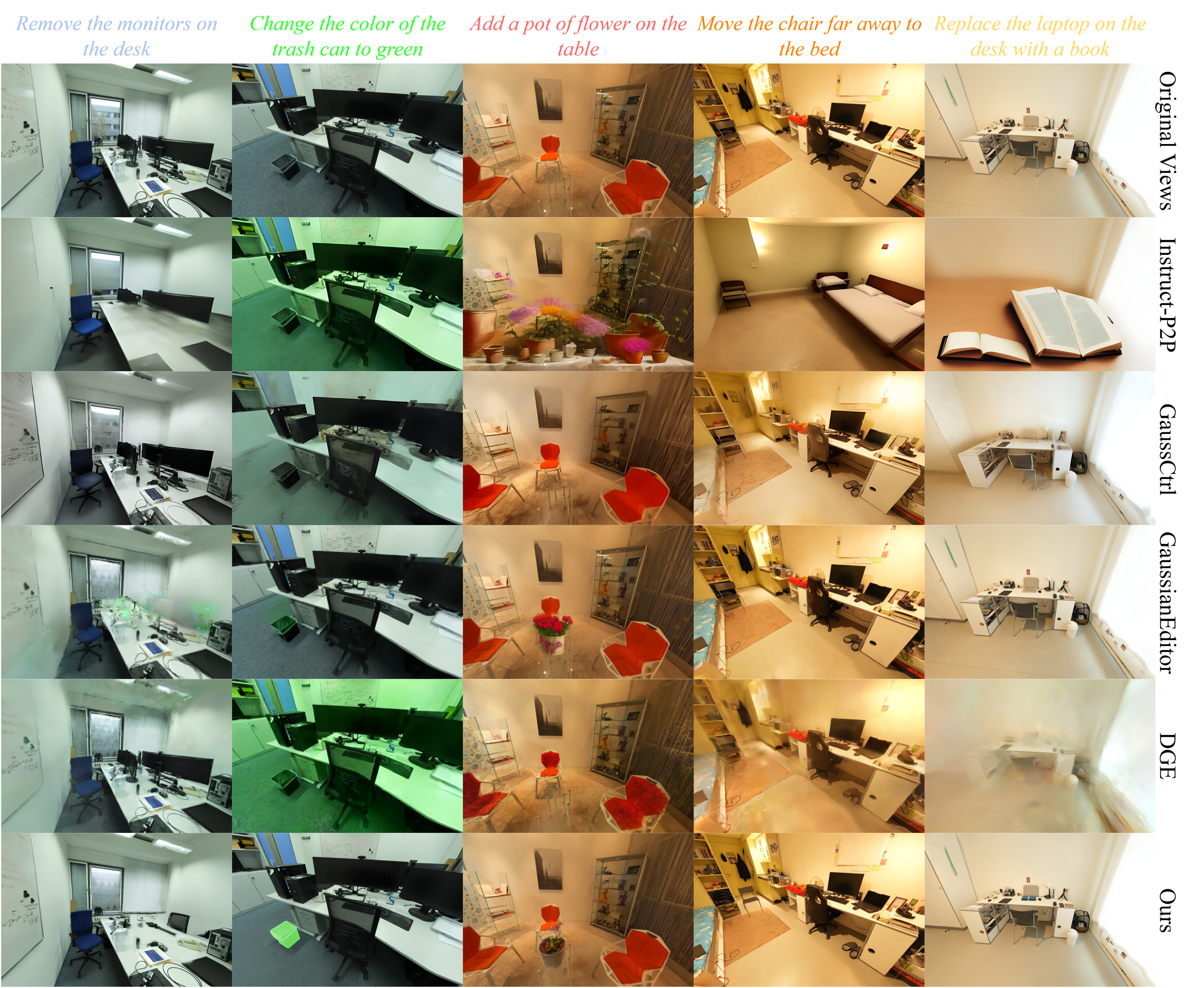}
   \caption{\textbf{Qualitative comparison with respect to other controllable scene editing approaches.} The experimental results show that in complex scene layouts, our 3D-only approach effectively utilizes 3D Gaussian spatial information, maintaining high editing quality independent of input image pixels and consistently preserving scene style, enabling more reliable and controllable editing in intricate settings.}
   \label{fig:Comparison Results}
    \vspace{-1.0em}
 \end{figure*}
 
 \textbf{Comparisons with other Gaussian-based controllable editing approaches.} Fig. \ref{fig:Comparison Results} reports a comparison of the performances of the proposed method with respect to GaussCtrl \cite{wu2024gaussctrl}, GaussianEditor \cite{chen2024gaussianeditor} and DGE \cite{chen2024dge}. Since all of them are based on Instruct-Pix2Pix \cite{brooks2023instructpix2pix}, which performs best at a resolution of 512x512 px, input images are resized to 512x512 px and then apply all types of edit operations supported by our pipeline across all methods. The results demonstrate that: 

(i) Our 3D-only pipeline uniquely provides stable, high-quality, and controllable edits in complex scenes, highlighting its distinct advantages over compared approaches. In contrast, the images generated by GaussCtrl and DGE show significant degradation in tasks involving object addition, recoloration movement, and replacement. This occurs because their pipelines process each frame individually with Instruct-Pix2Pix, which fails to maintain the original scene style and produces disrupted Gaussian features. While GaussianEditor uses Gaussian semantic tracking to define editable areas within the ROI, limitations in the diffusion model and semantic projection precision hinder its ability to deliver high-quality editing in complicated scenes. To further investigate this issue, we additionally process the same images directly with Instruct-Pix2Pix and show them also in Fig. \ref{fig:Comparison Results} with other approaches together. These outcomes further validate our previous observations.

(ii) Our approach uniquely supports object movement and replacement, as none of the compared methods effectively utilize the 3D spatial information of Gaussians. This demonstrates our pipeline’s ability to offer a wider range of editing operations.

\subsection{Quantitative Evaluation}
Table \ref{Table：Quantitative Evaluation} presents quantitative comparisons among all evaluated approaches, including ours. The metrics used are CLIP Text-Image Similarity (CTIS), CLIP Image-Image Similarity (CIIS), Running Time, and video RAM (VRAM) usage,  as measured based on the results in Fig. \ref{fig:Comparison Results}. 3DSceneEditor achieves the highest performance in both CTIS and IIS, indicating that our 3D-only design can better preserve scene style consistency while effectively responding to text instructions. For Running Time, 3DSceneEditor requires only 2-5 minutes for the initial edit\footnote[1]{Initial edit: Initialize the scene semantics with instance segmentation for object grounding and edit.} and less than 1 minute for secondary edits \footnote[2]{Secondary edit: Using saved instance semantics for object grounding and edit.}, significantly outperforming other approaches and enabling real-time scene editing. Additionally, our pipeline consumes slightly less GPU memory (VRAM) compared to others.

\begin{table*}[!tbp]
\caption{\textbf{Quantitative Comparison.} 3DSceneEditor surpasses other methods in both CLIP Text-Image Direction Similarity and Image-Image Similarity, requiring less time and GPU memory for scene editing, underscoring the advantages of our 3D-only architecture. In particular, our pipeline enables real-time editing for secondary edits, completing within 1 minute.}
\label{Table：Quantitative Evaluation}
    \centering
    
\begin{tabularx}{\textwidth}{c *{5}{Y}}
    \hline

        \textbf{Method}& \textbf{CLIP Text-Image Similarity (\%)}$\uparrow$ & \textbf{ CLIP Image-Image Similarity (\%)}$\uparrow$ & \textbf{Running Time}$\downarrow$ & \textbf{VRAM}$\downarrow$  \\ \hline
        GaussCtrl\cite{wu2024gaussctrl} & 22.01  & 91.90 & 8 min & 24000MB  \\ 
        GaussianEditor\cite{chen2024gaussianeditor} & 23.17 & 95.00& 8 min & 12000MB  \\ 
        DGE\cite{chen2024dge} & 22.96 & 86.42 & 4 min & 10000MB \\
        Ours & \textbf{23.20} & \textbf{96.17} & \textbf{2-5 min\footnotemark[1]  / $\leq$\textbf{}1 min\footnotemark[2]} & \textbf{9400MB\footnotemark[1] / \textbf{9100MB\footnotemark[2]}} \\ 

         \hline
    \end{tabularx}
 \vspace{-1.0em}
\end{table*}

\subsection{Ablation Study and Analysis}
\begin{figure}[h]
   \centering
   \includegraphics[width=\linewidth]{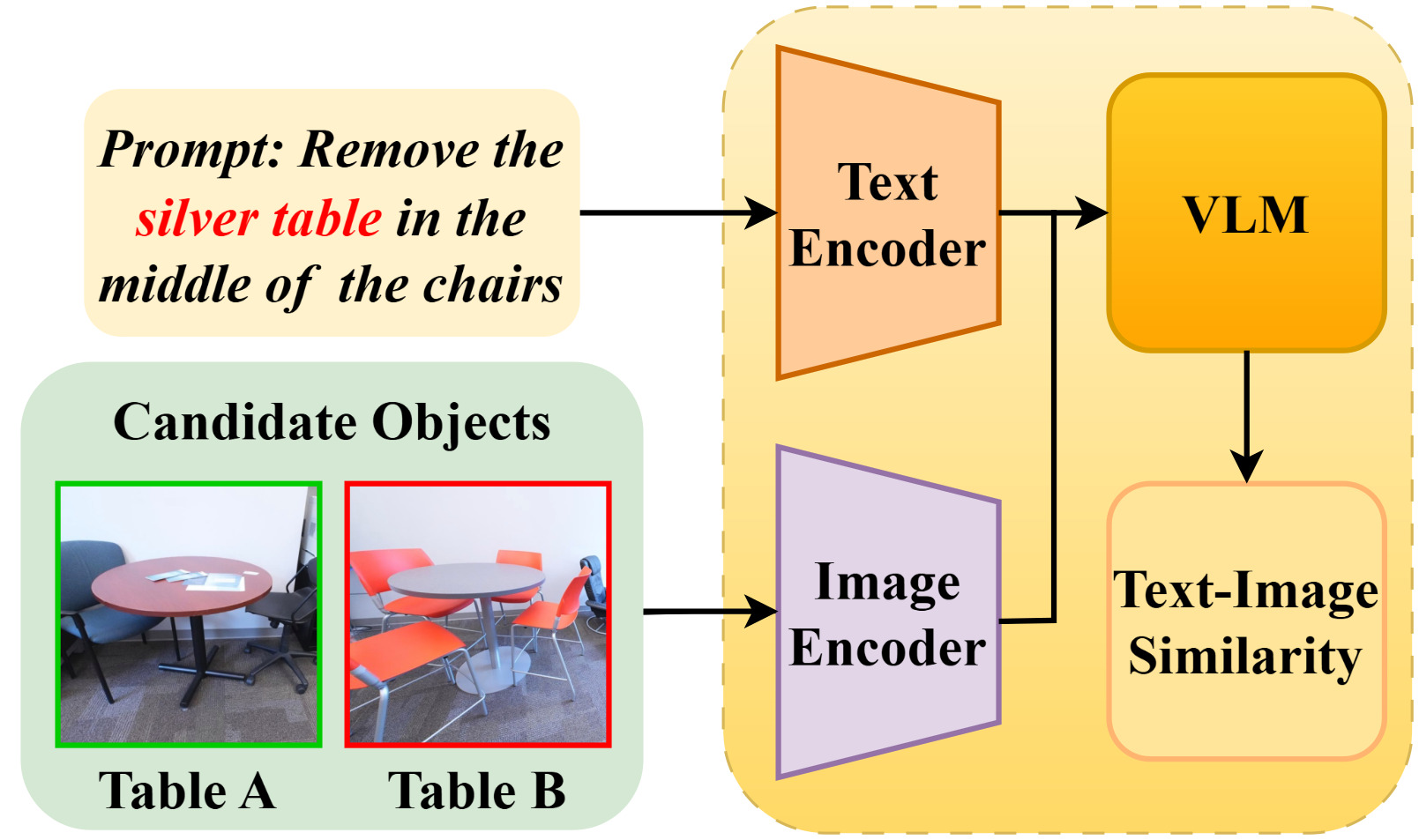}
   \caption{\textbf{Visualization of  Language-Object Correlation.} The keyword in the prompt (red font) is extracted and tokenized by a text encoder, then input into the vision-language model (VLM) alongside the image token. The Text-Image Similarity (TIS) is then calculated from the VLM's output.}
   \label{fig:ablation_llm}
 \end{figure}

\textbf{Ablation of language-object correlation.}
Table \ref{Table:ablation_llm} and Fig. \ref{fig:ablation_llm} illustrate the scalability of our Language-Object Correlation module with different Vision-Language Models (VLMs). Since the two tables in Fig. \ref{fig:ablation_llm} fit the description of "in the middle of the chairs," we further compare the results with various image-text encoders and show their Text-Image Similarity (TIS) in Table \ref{Table:ablation_llm}. With diverse encoder combinations, our module consistently makes correct decisions, demonstrating high flexibility and portability, allowing developers to interchange text and image encoders as needed.

\begin{table}[htbp]
    \centering
    \begin{threeparttable}[b]
    \begin{tabularx}{237pt}{c*{3}{Y}}
    \hline
    \textbf{\makecell{Encoder \\ (Image-Text)}}& \textbf{\makecell{Table A \\(\%)}}  & \textbf{\makecell{Table B \\ (\%)}} \\\hline
    CLIP-CLIP\cite{radford2021learning} & 24.25
  & \textbf{25.31} \\
    CLIP-Llama\cite{touvron2023llama} & 12.56
  &  \textbf{13.79}  \\
    Blip2-Blip2\cite{li2023blip} & 26.9 & \textbf{27.20}  \\
    CLIP-Qwen\cite{Qwen2VL} &15.23 & \textbf{17.31}  \\
    CLIP-BERT\cite{DBLP:journals/corr/abs-1810-04805} &23.94 & \textbf{24.16}  \\
    CLIP-Llama 2\tnote{1}\cite{touvron2023llama2} & 13.69 & \textbf{15.23}
    \\
  \hline
    \end{tabularx}
    \begin{tablenotes}
    \footnotesize
     \item[1] Llama 2 is compressed to FP16/INT4 by GWQ \cite{shao2024gwq}
   \end{tablenotes}
   \end{threeparttable}
       \caption{\textbf{Ablation study of Language-Object Correlation.} This table shows the TIS of the prompt and the two tables in Fig. \ref{fig:ablation_llm} using different combinations of image-text encoder. The object with the highest TIS is selected as the final choice.}
    \label{Table:ablation_llm}
    \vspace{-1.0em}
\end{table}

\begin{figure}[h]
   \centering
   \includegraphics[width=\linewidth]{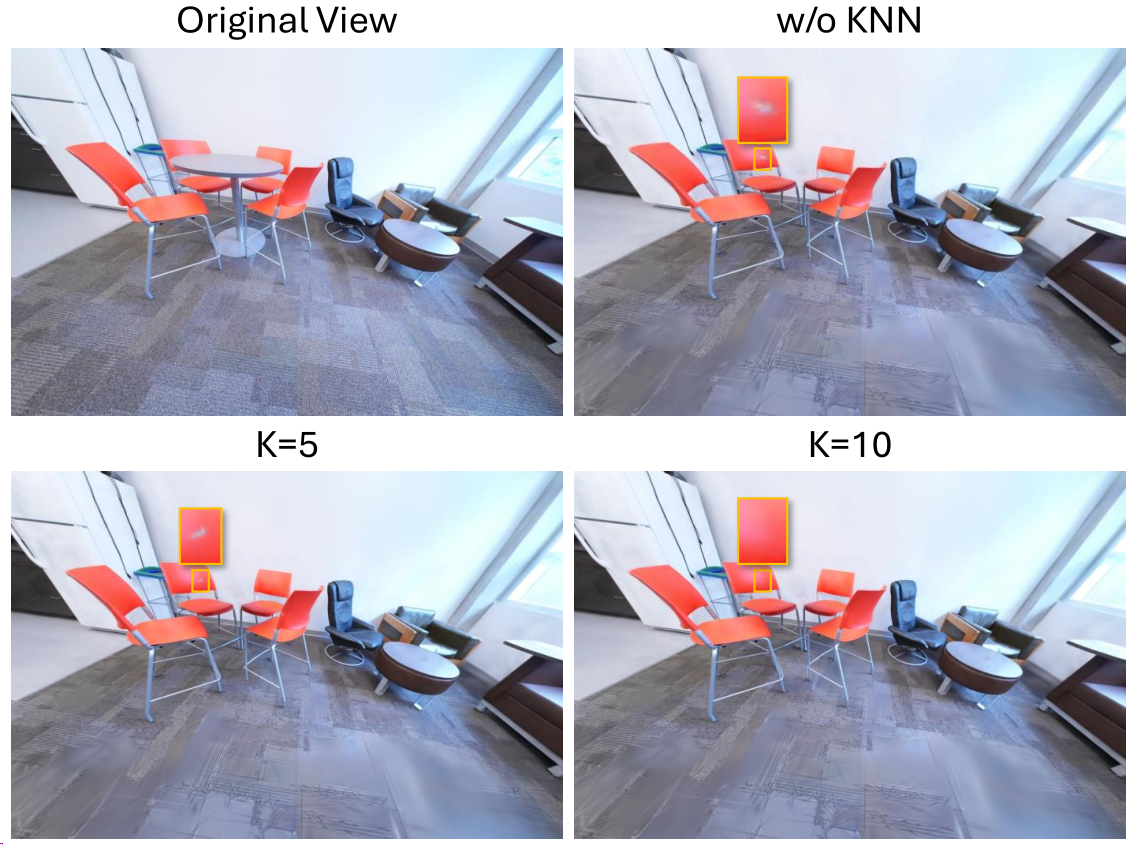}
   \caption{\textbf{Visualization of KNN in our optimizion module.} By using the KNN to refine the segmentation labels within ROI regions, we effectively reduces noise, and yielding more accurate editing.}
   \label{fig:ablation_knn}
   \vspace{-1.0em}
 \end{figure}

\textbf{Ablation of editing optimization.} Fig. \ref{fig:ablation_knn} reports ablation experiments on the Editing Optimization module. Without this module, edited images exhibit noise within the ROI, primarily due to segmentation errors near the junction between the chair and table. By adjusting the parameter $K$, our optimization module effectively reduces segmentation errors and enhances editing precision.

\label{Optimization of Inpainting}

\section{Conclusions}
This paper introduced 3DSceneEditor, an innovative 3D-only paradigm for text-guided, precise scene editing. To our knowledge, 3DSceneEditor is the first fully 3D-based approach for editing 3D Gaussians, fully leveraging 3D spatial information in Gaussians to enhance both efficiency and accuracy. Key techniques include applying instance segmentation to 3D Gaussians, extracting key instructions from the prompt, grounding the ROI to 3D Gaussians with a zero-shot object grounding module, and editing the scene within the Gaussian ROI. Our experiments demonstrated that 3DSceneEditor outperformed GaussCtrl \cite{wu2024gaussctrl}, GaussianEditor \cite{chen2024gaussianeditor} and DGE \cite{chen2024dge} with higher CTIS and CIIS scores, reduced running time, and lower GPU memory usage, validating its ability to achieve accurate, controllable, and real-time scene editing.
\section{Limitation and Future work}
In this paper, we focused on testing our paradigm in indoor scene editing, as the employed pretrained instance segmentation model was trained on the indoor scene dataset. In future work, we plan to explore and validate its effectiveness in more complex scenes. While our method addresses some issues inherited from integrated submodules, such as the misclassification around the intersection area between two instances, removing or moving certain Gaussians can still disrupt image rendering (e.g., impacting the color and texture of background areas overlapping the edited foreground across viewpoints). As mentioned in Section \ref{section:3.3}, in 3D Gaussian representation, a single Gaussian does not represent only an object exclusively.

{
    \small
    \bibliographystyle{ieeenat_fullname}
    \bibliography{main}
}

\clearpage
\setcounter{page}{1}
\maketitlesupplementary

\section{Implementation details}
\label{sec:Implementation details}
\subsection{Implementation Details of 3D Scene Representation}
\begin{figure}[h]
   \centering
   \includegraphics[width=\linewidth]{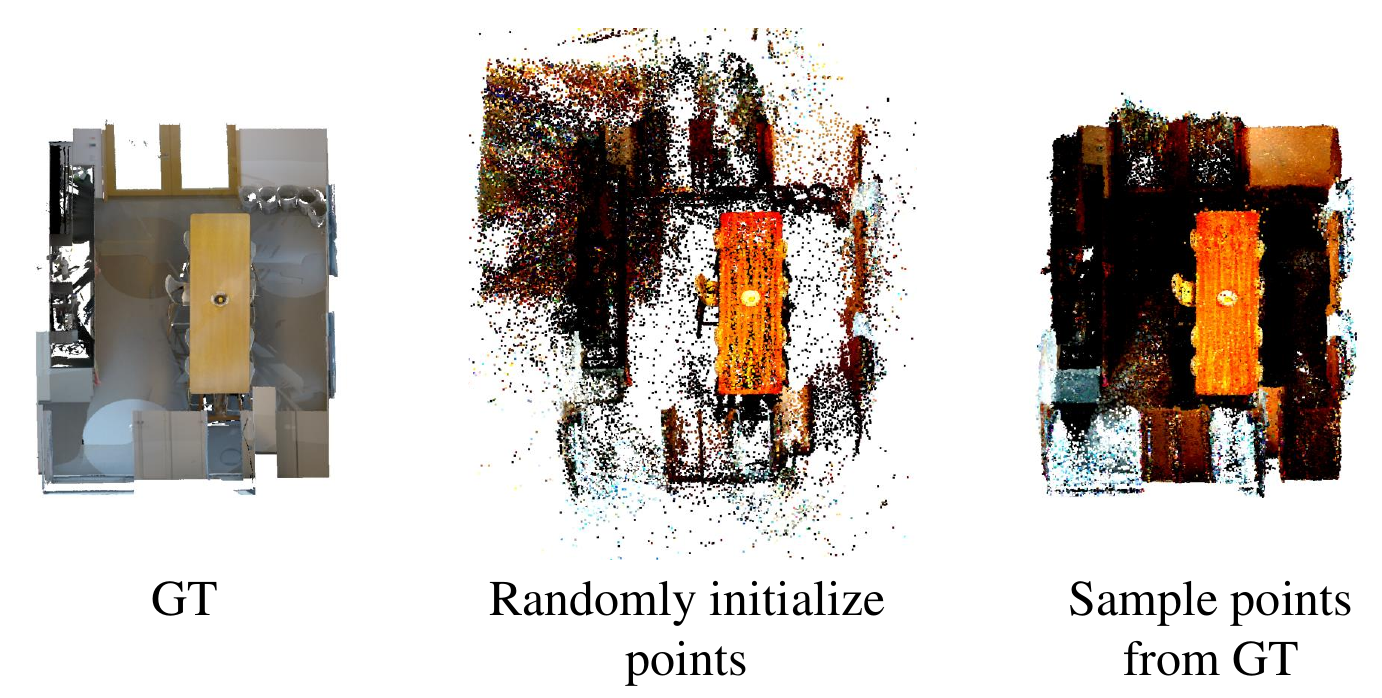}
       \vspace{-2.0em}
   \caption{\textbf{Visualization of geometric results for Gaussian sets with varying initializations.} The figure illustrates the results of using randomly initialized points versus points sampled from the ground truth (GT). The Gaussian set trained with random initialization shows significant noise in the 3D geometry, making it unsuitable for instance segmentation tasks. In contrast, the Gaussian set sampled from the GT and then trained achieves much higher geometric accuracy with minimal noise, closely aligning with the GT.}
   \label{fig:point initialization}
    \vspace{-1.0em}
 \end{figure}

3DSceneEditor processes 3D scenes reconstructed using 3D Gaussian Splatting \cite{kerbl20233d}. Since ScanNet++ \cite{yeshwanth2023scannet++} does not provide an initial point cloud derived from Structure-from-Motion (SfM)—a crucial requirement for achieving high-quality results with 3D Gaussian Splatting—we sampled 1 million points from the ground-truth (GT) mesh as the initial point cloud during training \cite{foroutan2024does}. This ensures the geometry of the Gaussian sets are well-defined (shown in Fig. \ref{fig:point initialization}), which is critical for subsequent instance segmentation tasks. 

Following the default configuration of 3D Gaussian Splatting, each scene is trained for 30,000 iterations, with input images exceeding 1600 pixels in width being automatically resized to 1600 pixels for computational efficiency.

For baseline methods, which utilize Instruct-Pix2Pix \cite{brooks2023instructpix2pix} for scene editing, input images are resized to 512×512 pixels as required, while all other hyperparameters are kept consistent.

\subsection{Implementation Details of Object Grounding}
\textbf{Instance segmentation.} In our experiments, we set the default confidence threshold  $c$ = 0.8 for instance segmentation to achieve higher segmentation precision. However, to avoid excluding small objects (e.g., paper, cups, books), we lower the threshold to $c$ = 0.3 when targeting such challenging objects for the pre-trained model, even if this results in additional noise or slight mis-segmentation. 

\begin{figure}[h]
   \centering
   \includegraphics[width=\linewidth]{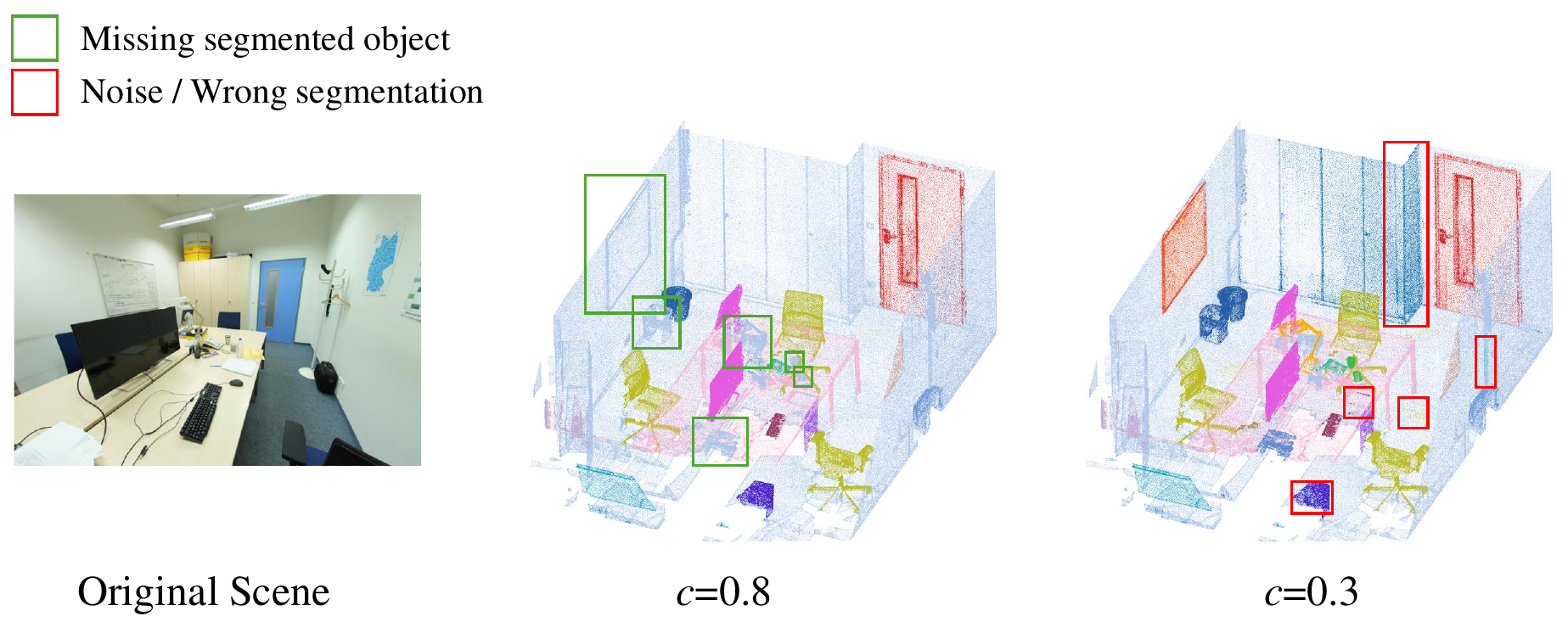}
   \caption{\textbf{Visualization of instance segmentation with different confidence threshold.} In this figure, we visualize instance segmentation of a 3D scene represented by a Gaussian set using colorful point clouds as Gaussians lack RGB attributes. Green bounding boxes highlight the missed segmented objects when the confidence threshold is set to $c$ = 0.8 compared to $c$ = 0.3, while red bounding boxes indicate additional noise or mis-segmentations introduced at the lower threshold.}
   \label{fig:instance segmentation confidence}
    \vspace{-1.0em}
 \end{figure}

In Fig. \ref{fig:instance segmentation confidence}, when the confidence threshold $c$ is set to 0.8, each instance is segmented more completely. However, objects with a confidence below $c$ are filtered out (highlighted by green bounding boxes). Conversely, at $c$ = 0.3, more objects are successfully segmented, but additional noise appears in some instances and on the floor (highlighted by red bounding boxes).

\textbf{Key words extract.} The prompts for our pipeline must begin with an operation keyword ( "remove," "add," "change," "move," or "replace"), followed by keywords for the target object, reference object, and reference direction. Keywords appearing before the reference direction are treated as the target object, while those after are considered reference objects. If no object keywords are detected after the reference direction, the object grounding module bypasses spatial relation interpretation. Table \ref{Table:key words extract} lists all keywords supported by our pipeline.

\begin{table}[!ht]
    \centering
    \begin{tabular}{m{3cm}<{\centering}|m{4.5cm}<{\centering}}
    \hline
        Operation & remove, add, change, move, replace   \\ \hline
        Target / Reference Object & open vocabulary  \\ \hline
        Reference Direction & left, right, middle, above, under, front, below, on, back, far away, close  \\ 
        \hline
        Color & refer to the color name in file "color-mapping.pdf" 
        \\ 
        \hline
    \end{tabular}
    \caption{Supported key words of 3DSceneEditor pipeline}
    \label{Table:key words extract}
\end{table}

\begin{figure*}[h]
   \centering
   \includegraphics[width=\linewidth]{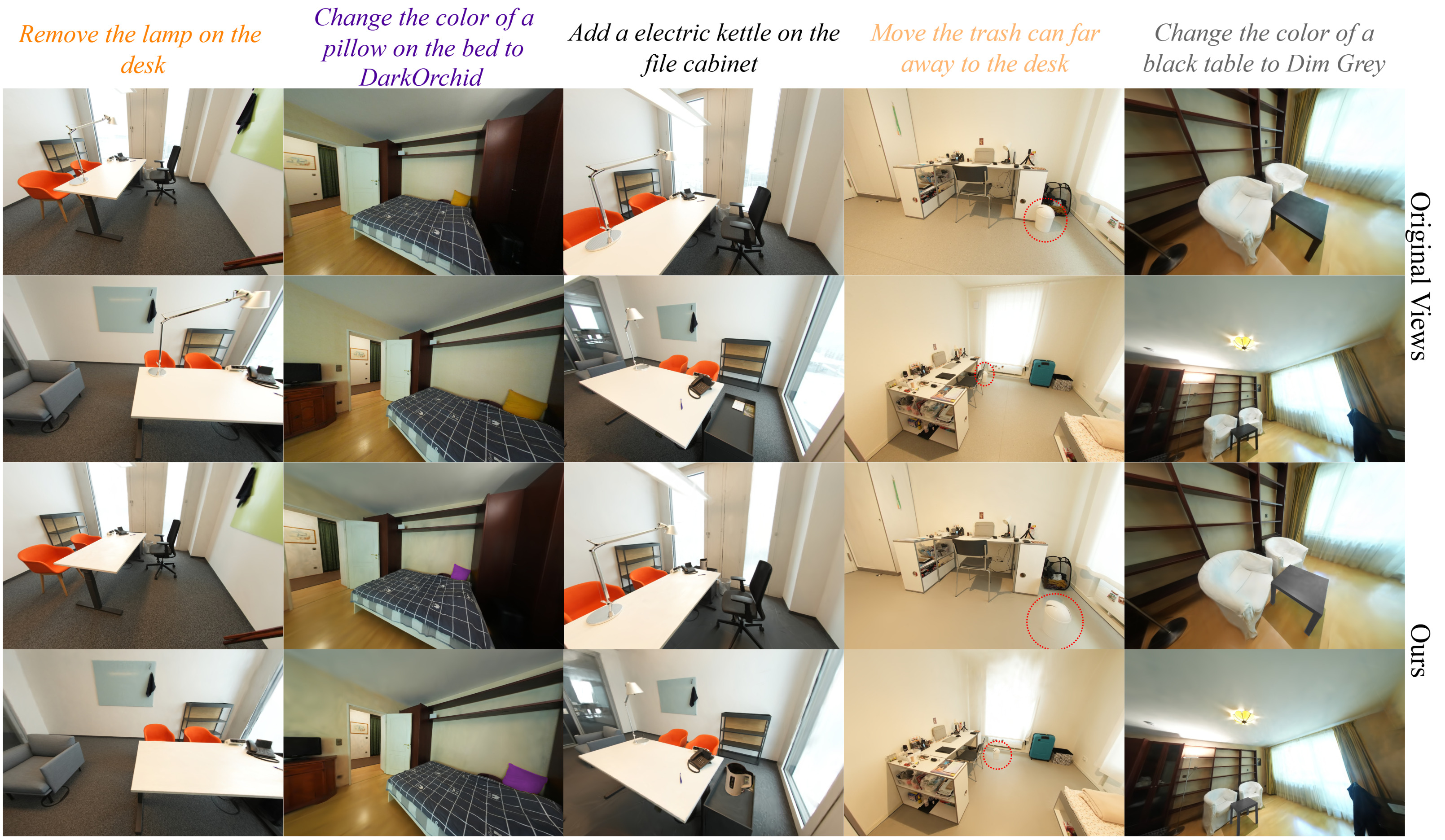}
   \caption{\textbf{More visualization result of 3DSceneEditor.} }
   \label{fig:experiment_result_sup}
    \vspace{-1.0em}
 \end{figure*}

\subsection{Implementation Details of 3D Gaussians Editing}
\textbf{Object re-coloration.} We designed a color-mapping table (refer to color-mapping.pdf) to translate color keywords from prompts into their corresponding RGB values. This pipeline enables editing with over 200 distinct colors, ensuring precise and flexible color adjustments.

\textbf{Object addtion and replacement.}
Our pipeline leverages DreamGaussian \cite{tang2023dreamgaussian} as the generative model for creating new objects. To ensure compatibility with 3D Gaussian Splatting, we set the spherical harmonics degree $sh$ = 3, while keeping the remaining parameters unchanged. The pipeline supports inputs in the form of text-only, image-only or text + image combinations for the generative model, with each generation trained for 500 epochs. 


\end{document}